\documentclass[letterpaper, 10 pt, conference]{ieeeconf}  

\usepackage[T1]{fontenc}
\usepackage[utf8]{inputenc}
\usepackage{authblk}
\usepackage{graphicx}
\usepackage[spaces,hyphens]{url}
\usepackage[linesnumbered,ruled]{algorithm2e}
\usepackage{array}
\usepackage{mathtools}
\usepackage{subcaption}
\usepackage{amssymb}
\usepackage{amsmath}
\usepackage{tikz}
\usepackage{diagbox}
\usepackage{fourier} 
\usepackage{flushend}
\usepackage[export]{adjustbox}
\usepackage{array}
\DeclareMathAlphabet{\mathcal}{OMS}{cmsy}{m}{n}
\usepackage{makecell}
\usepackage{stackengine}
\usepackage{multirow}
\usepackage[symbol]{footmisc}
\usepackage{sidecap}
\usepackage{tablefootnote}
\usepackage[flushleft]{threeparttable}
\usepackage{booktabs,caption}
\usepackage{xcolor, soul}
\usepackage{hyperref}
\sethlcolor{yellow}

\newcommand\normx[1]{\left\Vert#1\right\Vert}

\DeclareMathOperator*{\argmin}{argmin}

\IEEEoverridecommandlockouts                              

\title{\LARGE \bf
Active Pose Refinement for Textureless Shiny Objects using the Structured Light Camera
}

\author{Jun Yang*, Jian Yao$\dagger$, and Steven L. Waslander*
\thanks{This work was supported by Epson Canada Ltd.}
\thanks{*Jun Yang and Steven L. Waslander are with University of Toronto Institute for Aerospace Studies and Robotics Institute.
        {\tt\footnotesize \{jun.yang, steven.waslander\}@robotics.utias.utoronto.ca}}
\thanks{$\dagger$Jian Yao is with Epson Canada
        {\tt\footnotesize jian.yao@ea.epson.com}}
}

\begin{document}

\maketitle
\thispagestyle{empty}
\pagestyle{empty}

\begin{abstract}
6D pose estimation of textureless shiny objects has become an essential problem in many robotic applications. Many pose estimators require high-quality depth data, often measured by structured light cameras. However, when objects have shiny surfaces (e.g., metal parts), these cameras fail to sense complete depths from a single viewpoint due to the specular reflection, resulting in a significant drop in the final pose accuracy. To mitigate this issue, we present a complete active vision framework for 6D object pose refinement and next-best-view prediction. Specifically, we first develop an optimization-based pose refinement module for the structured light camera. Our system then selects the next best camera viewpoint to collect depth measurements by minimizing the predicted uncertainty of the object pose. Compared to previous approaches, we additionally predict measurement uncertainties of future viewpoints by online rendering, which significantly improves the next-best-view prediction performance. We test our approach on the challenging real-world ROBI dataset. The results demonstrate that our pose refinement method outperforms the traditional ICP-based approach when given the same input depth data, and our next-best-view strategy can achieve high object pose accuracy with significantly fewer viewpoints than the heuristic-based policies.
\end{abstract}

\section{INTRODUCTION}
Textureless shiny objects, such as metal parts, are essential components of many products. Detecting and estimating the poses of these objects is an important task in many robotic applications, such as bin-picking. Recently, with the explosive growth of deep learning techniques, many RGB-based solutions have been developed to address the pose estimation problem~\cite{xiang2018posecnn,sundermeyer2018implicit,deng2021poserbpf}. Although these approaches have shown good performance when projecting onto the 2D space, the actual 6D pose accuracy is still low due to the inherent scale and perspective ambiguities. Hence, in many object pose estimation systems, depth data is required to refine the pose accuracy.

\begin{figure}[t]
\centering
\begin{subfigure}{0.157\textwidth}
  \includegraphics[width=\linewidth]{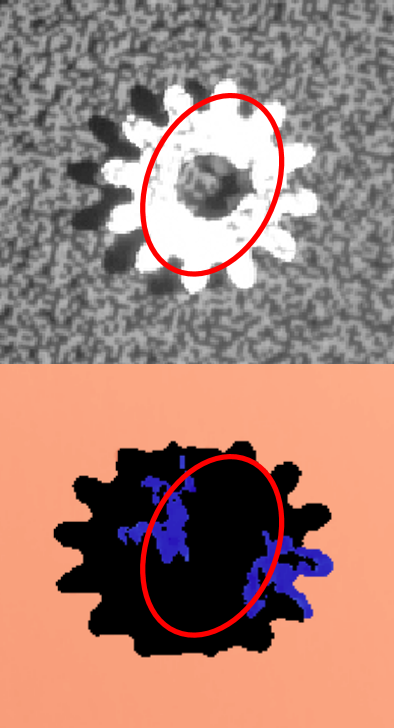}
  \vspace{-1.4\baselineskip}
  \caption{Image Saturation}
  \vspace{-0.5\baselineskip}
  \label{fig_shiny_GEAKS}
\end{subfigure}
\begin{subfigure}{0.157\textwidth}
    \includegraphics[width=\linewidth]{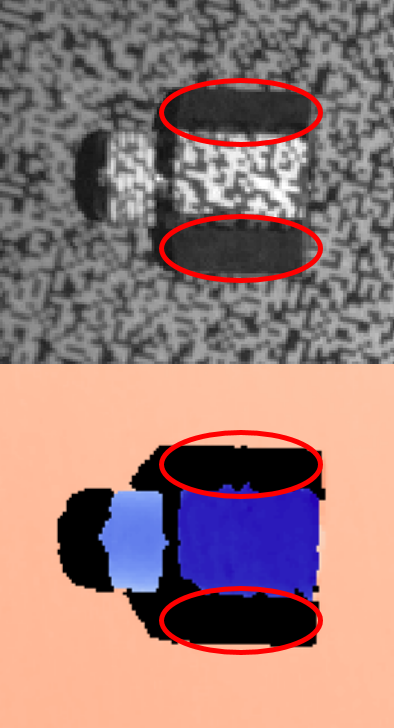}
    \vspace{-1.4\baselineskip}
    \caption{Low SNR}
    \vspace{-0.5\baselineskip}
    \label{fig_shiny_PF02}
\end{subfigure}
\begin{subfigure}{0.157\textwidth}
    \includegraphics[width=\linewidth]{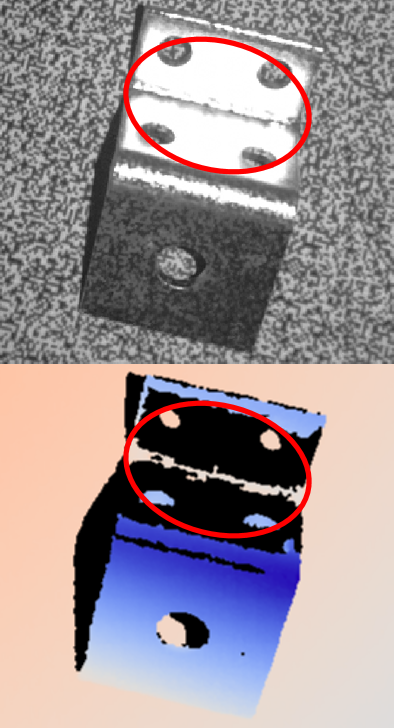}
    \vspace{-1.4\baselineskip}
    \caption{Inter-reflection}
    \vspace{-0.5\baselineskip}
    \label{fig_shiny_ZIGZAG}
\end{subfigure}
\caption{Missing depth measurements on shiny objects' surfaces using a structured light camera.}
\vspace{-1.2\baselineskip}
\label{fig_shiny}
\end{figure}

To acquire reliable depth maps for pose refinement, the structured light illumination (SLI) camera is usually used because of its high accuracy and resolution. It projects light patterns onto objects to simplify the stereo-matching problem and excels on diffuse surfaces. However, when imaging objects are highly reflective, the SLI camera produces depth maps with low accuracy and missing data. Due to specular reflection, a high proportion of the incident light is reflected, either directly back to the camera (image saturation), completely missing the camera (low SNR), or reflected within the object surfaces before returning to the camera (inter-reflection). As illustrated in Figure~\ref{fig_shiny}, each effect can result in inaccurate or missing depth measurements. To overcome this problem, our recent work fuses multi-view depth maps for higher levels of scene completion~\cite{yang2021probabilistic}. The remaining problems are selecting camera viewpoints to maximize information gain and utilizing the multi-view acquired depth for object pose estimation, which is crucial for fast scene understanding. 

Some approaches have been proposed that predict the next-best-view (NBV) to complete the depth data on the target objects and estimate/refine 6D object poses~\cite{wu2021object, yang2022next}. These studies assume that the complete depth data is necessary for the optimal object pose estimation, and aim to find camera viewpoints that can recover as much depth data as possible on all objects. However, this strategy is usually inefficient since, for the object pose estimation, the depth from some areas of the scene is far more important than others (e.g., those areas with lower measurement uncertainties and better constraints for the object pose refinement process). 

\begin{figure*}[t]
\centering
  \includegraphics[width=0.96\linewidth]{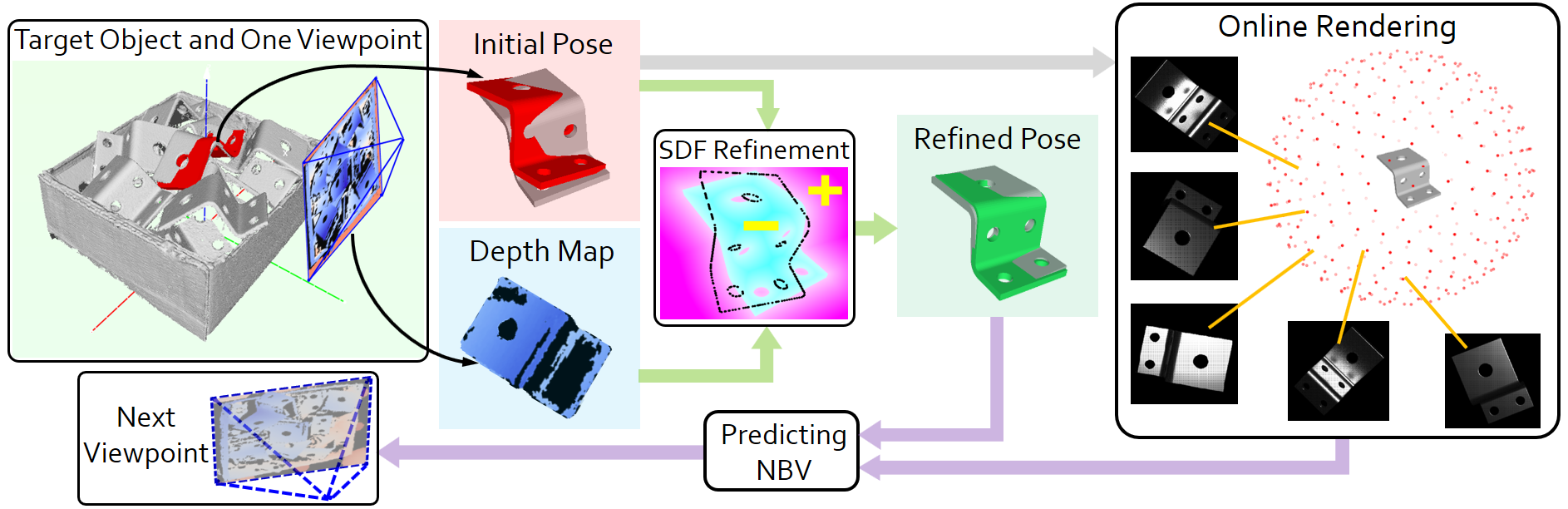}
\vspace{-0.4\baselineskip}  
\caption{An overview of the proposed multi-view pose refinement and the next-best-view prediction for the shiny objects.}
\label{fig_overview}
\vspace{-0.95\baselineskip}
\end{figure*}

The above observations motivated us to introduce a tightly coupled framework of 6D pose refinement and next-best-view prediction for textureless shiny objects. Inspired by~\cite{deng2021poserbpf,schmidt2014dart}, we first develop a signed distance function (SDF)-based optimization approach to refine the object pose. In addition, to mitigate the effect of depth errors, we estimate the depth uncertainties from the SLI camera and integrate them into our pose refinement module. Given the initial object pose, we iteratively refine it and determine the next best camera viewpoint to increase the pose accuracy. For the NBV prediction, our proposed method includes two main parts: a) a surface reflection model to predict the depth uncertainties, b) the NBV prediction for the object pose refinement by incorporating the reflection model. In the first part, we estimate the object's surface reflection parameters by differentiable rendering techniques. The estimated parameters are then used by online rendering to predict the object's depth uncertainties from a future viewpoint. In the second part, we integrate our reflection model into an information-theoretic NBV policy. For each candidate viewpoint, we predict the expected uncertainty of the object pose and determine the NBV by minimizing the predicted uncertainties. Figure~\ref{fig_overview} shows an overview of the framework. In the sections~\ref{sec_RF}-\ref{sec_NBV}, we will describe each part in detail.

We evaluate our framework on the challenging ROBI dataset~\cite{yang2021robi}. We first evaluate our pose refinement with passive viewpoint selection, showing that our refinement module outperforms the widely used iterative closest point (ICP) approach when given the same input depth measurements. To demonstrate the advantages of our NBV policy, we compare it with two heuristic-based strategies. The results indicate our method can achieve high pose refinement accuracy using significantly fewer viewpoints. In summary, our key contributions are:
\begin{itemize}
    \item A 6D pose refinement approach for textureless shiny objects designed for SLI cameras. Our approach comprises (a) the estimation of pixel depth uncertainties, (b) the integration of uncertainty estimates within our SDF-based object pose refinement module.
    \item A surface reflection model to predict the object's depth uncertainties for unseen camera viewpoints. Our reflection model recovers the object's reflection parameters with the differentiable renderer.
    \item An active vision system that integrates the reflection model and our pose refinement module via the online rendering to predict the next-best-view for active pose estimation.
\end{itemize}

\section{RELATED WORK}
\subsection{Object Pose Refinement}
To acquire highly accurate object poses, pose refinement is a critical step and has been mostly addressed using depth data. Iterative Closest Point (ICP) and its variants~\cite{rusinkiewicz2001efficient} are the most classical approaches and have been used in many object pose estimation pipelines~\cite{xiang2018posecnn,sundermeyer2018implicit}. Given the initial object pose, ICP refines it iteratively by establishing the point-to-point correspondences from the 3D point cloud to the object model and minimizing the distances. To improve the runtime performance, several approaches~\cite{deng2021poserbpf,schmidt2014dart,zhang2019fast} have been proposed to reduce the computation cost. Among them, Deng et al.~\cite{deng2021poserbpf} avoid the costly point correspondence building and refine the object pose by matching the 3D points from the depth measurements against the SDF of the target object. Other approaches improve pose refinement by replacing ICP with a neural network~\cite{wang2019densefusion,wen2020se}. The most representative work is DenseFusion~\cite{wang2019densefusion}, which fuses the RGB and depth features and trains a deep object pose refinement network to iteratively regress a pose offset.

\subsection{Depth Acquisition with Structured Light Camera} 
Structured Light Illumination (SLI) cameras are one of the most used indoor 3D sensors, but they produce inaccurate and missing depth measurements when target objects have shiny surfaces. To overcome the image saturation problem (Figure~\ref{fig_shiny_GEAKS}), high dynamic range (HDR)-based methods are widely used in many SLI systems~\cite{zhang2009high}. HDR methods fuse a set of images under multiple exposures into a single image for stereo matching. To reduce HDR's time cost, Liu et al.~\cite{liu2020optical} employed a neural network to directly enhance single exposure-captured images. Despite good performance, these methods cannot solve the low SNR and inter-reflection problems (illustrated in Figure~\ref{fig_shiny_PF02}~and~\ref{fig_shiny_ZIGZAG}) from a single viewpoint. In comparison, when the setup permits, multi-view acquisition~\cite{yang2021probabilistic,yang2022next} can provide a high level of depth completion.

\subsection{Active Vision}
Active vision \cite{aloimonos1988active} refers to actively manipulating the camera viewpoint to obtain the maximum information for different tasks. Active vision has received a lot of attention from the robotics community and has been employed in many applications, such as robot manipulation~\cite{morrison2019multi}, reconstruction~\cite{wu2021object,yang2022next,forster2014appearance} and SLAM~\cite{davison2002simultaneous,zhang2018perception,zhang2019beyond}. Recent studies show that active vision can be achieved by maximizing the Fisher information of the robot state~\cite{forster2014appearance,zhang2018perception,zhang2019beyond}. For example, the authors in~\cite{zhang2018perception,zhang2019beyond} use the Fisher information to find highly-informative trajectories and achieve high localization accuracy.

\section{MULTI-VIEW OBJECT POSE REFINEMENT}
\label{sec_RF}
This section presents our multi-view 6D pose refinement formulation for shiny objects with the SLI camera. Given the 3D object model and multi-view acquired depth maps, we aim to refine the rigid pose $\boldsymbol{T}_{ow} \in SE(3)$ from a global (world) coordinate ${W}$ to the object coordinate ${O}$. We assume that we know the camera poses $\boldsymbol{T}_{wc} \in SE(3)$ with respect to the world coordinate. Our pose refinement module consists of two parts: (a) depth uncertainty estimation from the SLI camera, (b) an optimization-based object pose refinement module that takes the uncertainty estimates into account. In the following subsections, we describe these two parts in detail.

\subsection{Estimating Measurement Uncertainty}
\label{sec_uncertainty}
For an SLI camera, the depth measurement uncertainty is a function of the depth, camera parameters (e.g., intrinsics), and the photometric appearance of the projected light patterns. In this section, we describe how to compute the depth uncertainty, starting from estimating the uncertainty for the disparity, $\boldsymbol{\sigma}_{d}^2$, and propagating it through a non-linear model to obtain the depth uncertainty, $\boldsymbol{\sigma}_{z}^2$.

Depending on the hardware design of an SLI camera, the stereo matching is performed from camera-to-camera or camera-to-projector. We use the camera-to-camera in our derivation, which can be easily adapted to the camera-to-projector design. Given the stereo pair of left $\boldsymbol{I}_L$ and right $\boldsymbol{I}_R$ pattern projected images, the disparity uncertainty, $\boldsymbol{\sigma}_{d}^2$, accounts for the appearance ambiguities between image patches. Intuitively, matching is reliable for image patches with strong image intensity gradients. When the dominant gradient direction is parallel to the epipolar line ($x$ axis for the left-right setup), the obtained disparity becomes more reliable, which results in lower uncertainty and vice versa. Inspired by~\cite{forster2014appearance}, we use the sum of squared differences (SSD) to quantify the disparity and define the photometric error as:
\begin{align}
\label{equ_ssd}
    e = \boldsymbol{I}_L\left(u_L, v \right) - \boldsymbol{I}_R\left(u_R, v\right)
    = \boldsymbol{I}_L\left(u_L, v \right) - \boldsymbol{I}_R\left(u_L-d, v\right)
\end{align}
where $d$ is the acquired disparity for the pixel $\mathbf{u} = \left[u,v\right]$ in the left image $\boldsymbol{I}_L$. We assume the disparity of a pixel is normally distributed and compute its variance, $\boldsymbol{\sigma}_{d}^2$, through the Fisher information: 
\begin{equation}
\label{equ_var_disparity}
    \boldsymbol{\sigma}_{d}^2 = \boldsymbol{\sigma}_I^2 \: {\left(\boldsymbol{\mathsf{J}}_d^T \: \boldsymbol{\mathsf{J}}_d\right)}^{-1}
\end{equation}
where $\boldsymbol{\sigma}_I^2$ denotes the variance of the image noise and Jacobian $\boldsymbol{\mathsf{J}}_d$ is derived by:
\begin{equation}
\label{equ_jac_disparity}
    \boldsymbol{\mathsf{J}}_d = \frac{\partial e}{\partial d} = -\frac{\partial \boldsymbol{I}_R}{\partial u_R} \frac{\partial u_R}{\partial d} = \frac{\partial \boldsymbol{I}_R}{\partial u_R}
\end{equation}
which is the image gradient along the x-axis over a patch from image $\boldsymbol{I}_R$, centered at the pixel $\boldsymbol{u}_R=\left[u_R, v\right]$. To obtain the measurement variance of the depth, $\boldsymbol{\sigma}_{z}^2$, we propagate the disparity variance, $\boldsymbol{\sigma}_{d}^2$, through:
\begin{equation}
\label{equ_var_depth}
    \boldsymbol{\sigma}_{z}^2 = \boldsymbol{F}\: \boldsymbol{\sigma}_{d}^2 \: \boldsymbol{F}^T
\end{equation}
where $\boldsymbol{F}$ is the Jacobian of depth, $\boldsymbol{z}$, with respect to disparity, $\boldsymbol{d}$. For the camera-to-camera setup, the acquired depth will have high uncertainty when the image gradient is weak in the left or right image. Hence, we need to compute $\boldsymbol{\sigma}_{z, L}^2$ and $\boldsymbol{\sigma}_{z, R}^2$ for $\boldsymbol{I}_L$ and $\boldsymbol{I}_R$, respectively, and final depth uncertainty is obtained by:
\begin{equation}
\label{equ_var_depth2}
    \boldsymbol{\sigma}_{z}^2 = \max \left( \boldsymbol{\sigma}_{z,L}^2 , \boldsymbol{\sigma}_{z,R}^2\right)
\end{equation}

\begin{figure}[t]
\centering
\begin{subfigure}{0.42\textwidth}
    \includegraphics[width=\linewidth]{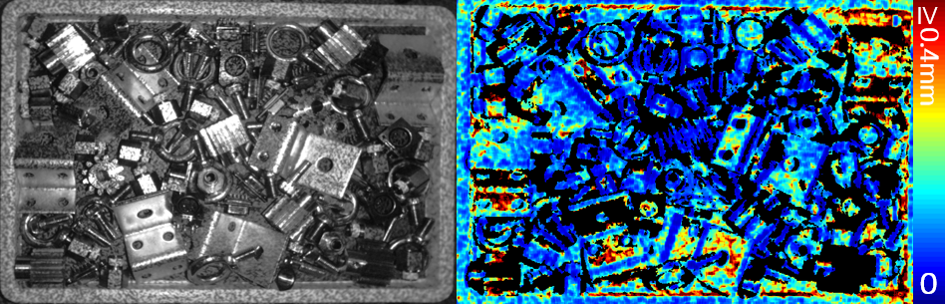}
    \vspace{-0.8\baselineskip}
    \label{fig2a}
\end{subfigure}
\begin{subfigure}{0.43\textwidth}
    \includegraphics[width=\linewidth]{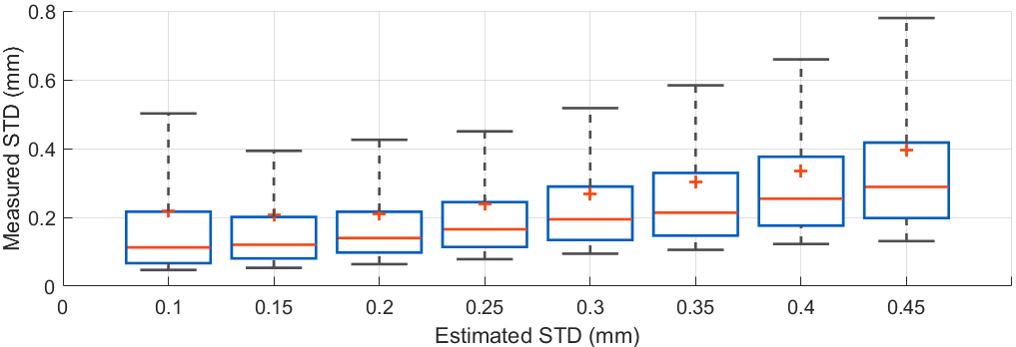}
    \vspace{-0.8\baselineskip}
    \label{fig2b}
\end{subfigure}
\vspace{-0.4\baselineskip}
\caption{Upper: The pattern projected image and the estimated uncertainties on the depth map. Lower: The correlation between estimated and measured uncertainty.}
\label{fig_unc}
\vspace{-0.8\baselineskip}
\end{figure}

We demonstrate our estimated uncertainty measure in Figure~\ref{fig_unc}. We can see that the estimated uncertainty accurately correlates with the actual measured depth variance. 

\subsection{Pose Refinement With SDFs}
\begin{figure}[t]
\centering
  \includegraphics[width=0.98\linewidth]{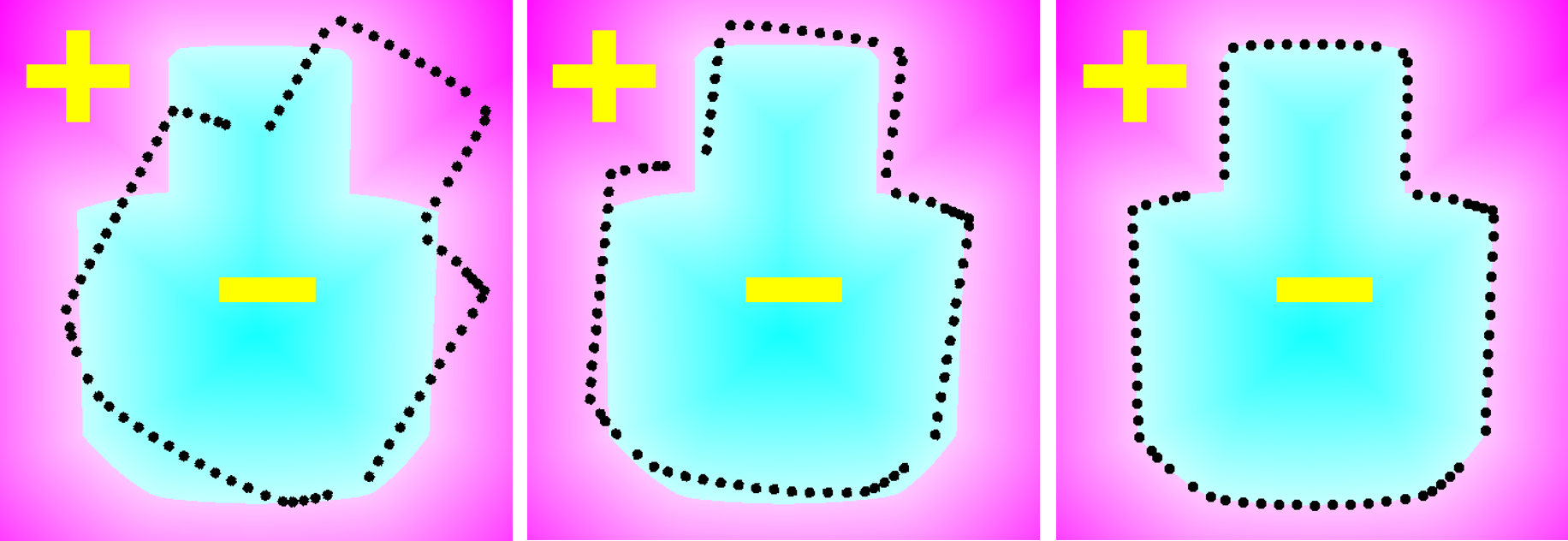}
  \caption{Object pose refinement module. Pink and blue represents the space with positive and negative distance, respectively. Black dots are the transformed point cloud. From left to right: the object pose, $\boldsymbol{T}_{ow}$, is refined iteratively by minimizing the SDF loss (Equation~\ref{equ_SDF}).}
  \vspace{-0.9\baselineskip}
  \label{fig_sdf}
\end{figure}

We formulate the object pose refinement as an optimization problem and solve it iteratively. Our refinement process is illustrated in Figure~\ref{fig_sdf}. Based on a signed distance function ($SDF$) approach~\cite{deng2021poserbpf,schmidt2014dart}, we refine the object pose $\boldsymbol{T}_{ow}$ by matching the depth measurement, $z$, which is defined for each pixel $\mathbf{u} = \left[u,v\right]$, against the $\boldsymbol{\mathsf{SDF}}$ of the target object model.

Given the depth map ${\boldsymbol{Z}_k(\boldsymbol{u})}$ from viewpoint $k$, we first extract the object mask, $\boldsymbol{M}_k$, and obtain the object's depth measurements. We utilize an instance segmentation network from~\cite{yang20236d} which provides pixel-level instance predictions. By back-projecting the pixels in $\boldsymbol{M}_k$, we obtain the point cloud of the object, $\boldsymbol{P}_{c,k} \in \mathbb{R}^3$, defined in the $k^{th}$ camera coordinate as:
\begin{equation}
\label{equ_obj}
    \boldsymbol{P}_{c,k} = \Bigl\{{\boldsymbol{Z}_k(\boldsymbol{u})} \: \boldsymbol{K}^{-1} \: \left[ \boldsymbol{u},1 \right]^T \:,\: \boldsymbol{u}\in\boldsymbol{M}_k \Bigl\}
\end{equation}
where $\boldsymbol{K}$ represents the camera intrinsic matrix. We transform the point cloud, $\boldsymbol{P}_{c,k}$, to the world coordinate frame, ${W}$, with the known camera pose, $\boldsymbol{T}_{wc,k}$:
\begin{equation}
\label{equ_obj}
    \boldsymbol{P}_{w} = \Bigl\{ \boldsymbol{T}_{wc,k} \: \boldsymbol{P}_{c,k} \:,\: k=1:K \Bigl\}
\end{equation}
where $\boldsymbol{P}_w$ is the point cloud defined in the world frame. We optimize the object pose $\boldsymbol{T}_{ow}$ by matching the 3D points against the $\boldsymbol{\mathsf{SDF}}$ of the target object model:
\begin{equation}
\label{equ_SDF}
    \boldsymbol{T}_{ow}^* = \argmin{\sum_{\boldsymbol{p}_{w,i} \in \boldsymbol{P}_{w}} {{\normx{\boldsymbol{\mathsf{SDF}} \left( \boldsymbol{T}_{ow} \: \boldsymbol{p}_{w,i} \right)}}^2}}
\end{equation}
where $\boldsymbol{p}_{w,i}$ is a 3D point in the point cloud $\boldsymbol{P}_{w}$. The function $\boldsymbol{\mathsf{SDF}}\left( \boldsymbol{T}_{ow}\: \boldsymbol{p}_{w,i}\right)$ denotes the signed distance value by transforming the 3D point $\boldsymbol{p}_{w,i}$ from the world frame to the object model frame with a pose estimate $\boldsymbol{T}_{ow}$.

To estimate the pose, $\boldsymbol{T}_{ow}$, from measurements $\boldsymbol{P}_{w}$, we formulate the problem as a nonlinear least squares (NLLS) problem. We solve this problem with the Gauss-Newton algorithm and integrate the $\boldsymbol{\mathsf{SDF}}$ measurement uncertainties, $\boldsymbol{\Sigma}_{\boldsymbol{\mathsf{sdf}}}$, in each iterative step:
\begin{equation}
\label{equ_GN}
    \left( \boldsymbol{\mathsf{J}}_{\boldsymbol{\xi}_{ow}}^T \: \boldsymbol{\Sigma}_{\boldsymbol{\mathsf{sdf}}}^{-1} \: \boldsymbol{\mathsf{J}}_{\boldsymbol{\xi}_{ow}} \right) \delta_{\boldsymbol{\xi}_{ow}} = \boldsymbol{\mathsf{J}}_{\boldsymbol{\xi}_{ow}}^T \boldsymbol{\Sigma}_{\boldsymbol{\mathsf{sdf}}}^{-1} \: \boldsymbol{\mathsf{SDF}}
\end{equation}
where $\boldsymbol{\mathsf{J}}_{\boldsymbol{\xi}_{ow}}$ is the stacked Jacobian matrix of $\boldsymbol{\mathsf{SDF}}$:
\begin{align}
\label{equ_jacobian_Tow}
    \boldsymbol{\mathsf{J}}_{\boldsymbol{\xi}_{ow}} = \frac{\partial \boldsymbol{\mathsf{SDF}}}{\partial \boldsymbol{\xi}_{ow}} = \frac{\partial \boldsymbol{\mathsf{SDF}}}{\partial \boldsymbol{P}_o}   \frac{\partial \boldsymbol{P}_o}{\partial \boldsymbol{\xi}_{ow}}
\end{align}
where $\boldsymbol{\xi}_{ow}\in \mathfrak{se}(3)$ is the Lie algebra representation of the transformation $\mathbf{T}_{ow}$, and $\boldsymbol{P}_o$ is the point cloud, transformed to the object frame. We acquire the uncertainty, $\boldsymbol{\Sigma}_{\boldsymbol{\mathsf{sdf}}}$, by propagating the depth uncertainty (obtained from Section~\ref{sec_uncertainty}), $\boldsymbol{\sigma}_{z}^2$, through a nonlinear model:
\begin{equation}
\label{equ_propagate}
    \boldsymbol{\Sigma}_{\boldsymbol{\mathsf{sdf}}} = \boldsymbol{G} \: \boldsymbol{\sigma}_{z}^2 \: \boldsymbol{G}^T
\end{equation}
where $\boldsymbol{G}$ is the Jacobian of $\boldsymbol{\mathsf{SDF}}$ value with respect to the depth measurement $\boldsymbol{z}$.

\section{Predicting Depth Uncertainty}
\label{sec_Predict_depth_uc}
The object pose refinement performance relies heavily on the input depth measurements from different viewpoints. For an SLI camera, to find the optimal viewpoint, it is important to quantify the depth uncertainty for future viewpoints. In this section, we detail how to predict the depth uncertainty by the rendering technique. The predicted uncertainties will be used to find the next-best-view for the object pose refinement (Section~\ref{sec_NBV}).

\subsection{Image Acquisition Process}
\label{sec_IP}
\begin{figure}[t]
\centering
  \includegraphics[width=\linewidth]{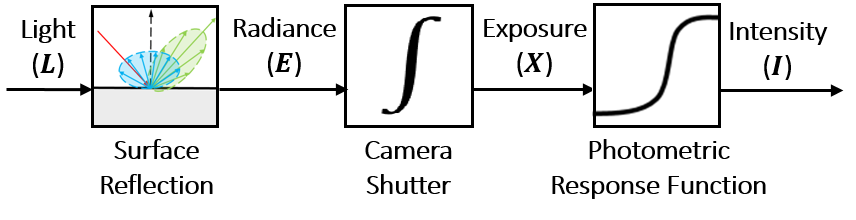}
  \vspace{-1.8\baselineskip}
  \caption{Image acquisition process of the SLI camera.}
  \vspace{-0.8\baselineskip}
\label{fig_imaging}
\end{figure}

The depth acquisition of an SLI camera is influenced by the light sources, camera viewpoint, and object characteristics (e.g., surface materials). Figure~\ref{fig_imaging} illustrates the image acquisition process of the SLI camera. Typically, two light sources need to be considered: the ambient light, $\boldsymbol{L_a}$, and the projector light, $\boldsymbol{L_p}$. Since the projector light is the dominating light source and the ambient light is negligible in comparison, we define the total light source $\boldsymbol{L_{total}}$ as:
\begin{equation}
\label{equ_light}
    \boldsymbol{L_{total}} = \boldsymbol{L_p} + \boldsymbol{L_a} \approx \boldsymbol{L_p}
\end{equation}

\begin{figure}[t]
\centering
  \includegraphics[width=0.9\linewidth]{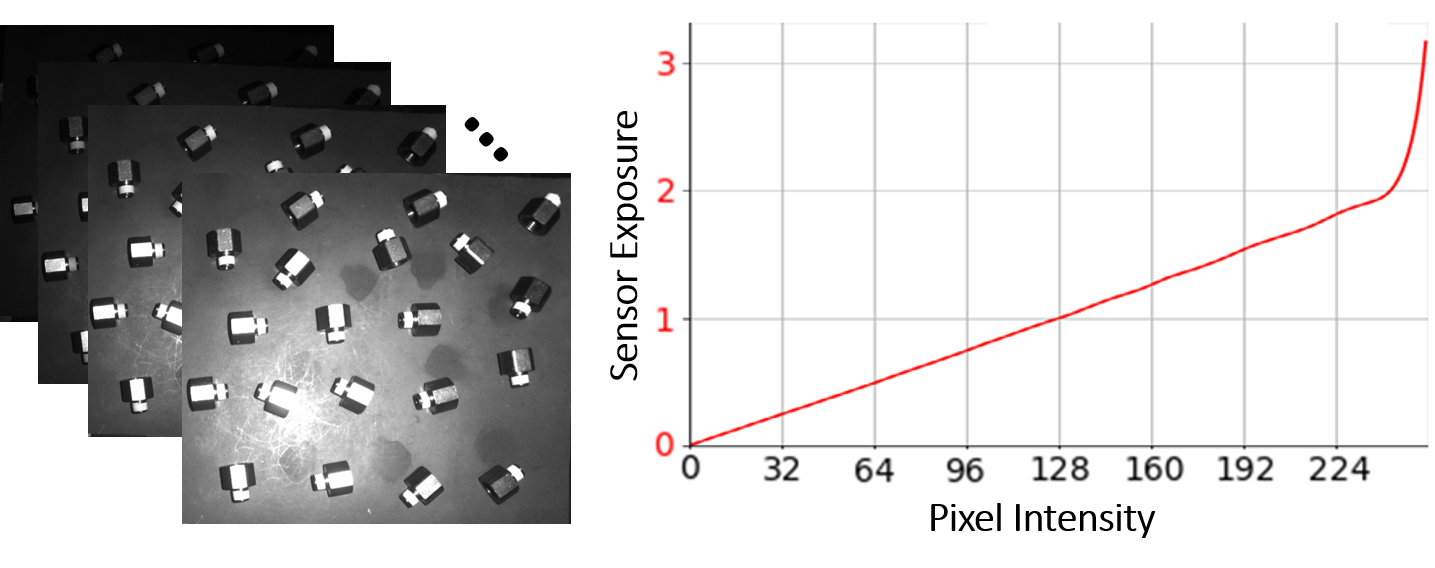}
  \vspace{-0.6\baselineskip}
  \caption{Left: Input images for calibrating the photometric response function. Right: The recovered function.}
  \vspace{-0.9\baselineskip}
\label{fig_photo}
\end{figure}

Given the light source and other scene parameters (e.g., object poses and materials), the reflection function, $\mathnormal{f}(\boldsymbol{\cdot})$, returns the radiance, $\boldsymbol{E}$, which is the amount of light that reflects into the camera lens per time unit. We recover the reflection function using a differentiable rendering algorithm. The details are described in Section~\ref{sec_DR}. The sensor exposure, $\boldsymbol{X}$, then integrates the received radiance, $\boldsymbol{E}$, within the camera exposure time, $\boldsymbol{\Delta t}$, via the camera shutter. The photometric response function, $g\left(\boldsymbol{X}\right)$, finally maps the exposure $\boldsymbol{X}$ to the pixel intensity $\boldsymbol{I}$ in the pattern projected image:
\begin{equation}
\label{equ_photometric}
    \boldsymbol{I} = g\left(\boldsymbol{X}\right)= g\left(\boldsymbol{E \Delta t}\right)
\end{equation}
We obtain the function $g\left(\boldsymbol{X}\right)$ and its inverse, $g^{-1}\left({\boldsymbol{I}}\right)$, using a photometric calibration approach, presented in~\cite{debevec2008recovering}. The input to the calibration process is a number of images taken from a static scene with different known exposures, $\boldsymbol{\Delta t}$. A white pattern is projected onto the scene during the capture. An example of input images and the recovered photometric response function, $g(\boldsymbol{X})$, is shown in Figure~\ref{fig_photo}.

\begin{figure}[t]
\centering
\begin{subfigure}{0.48\textwidth}
    \includegraphics[width=\linewidth]{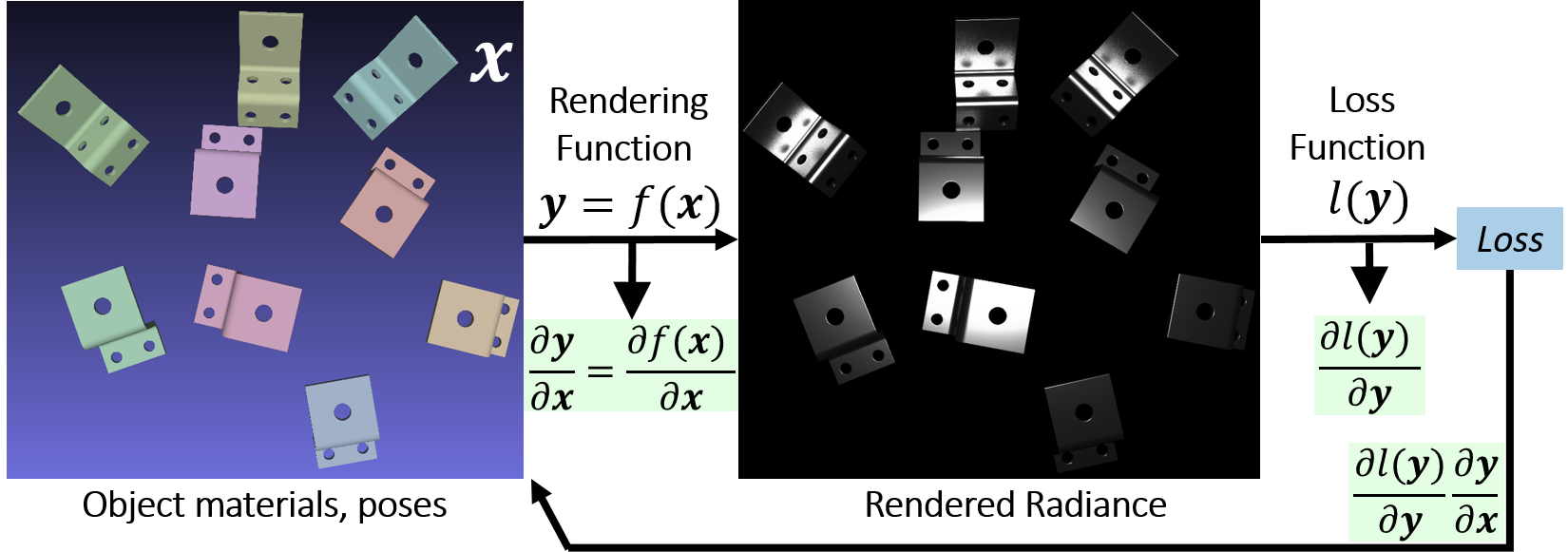}
    \vspace{-1.2\baselineskip}
    \caption{}
    \label{fig_dr1}
\end{subfigure}
\begin{subfigure}{0.26\textwidth}
    \includegraphics[width=\linewidth]{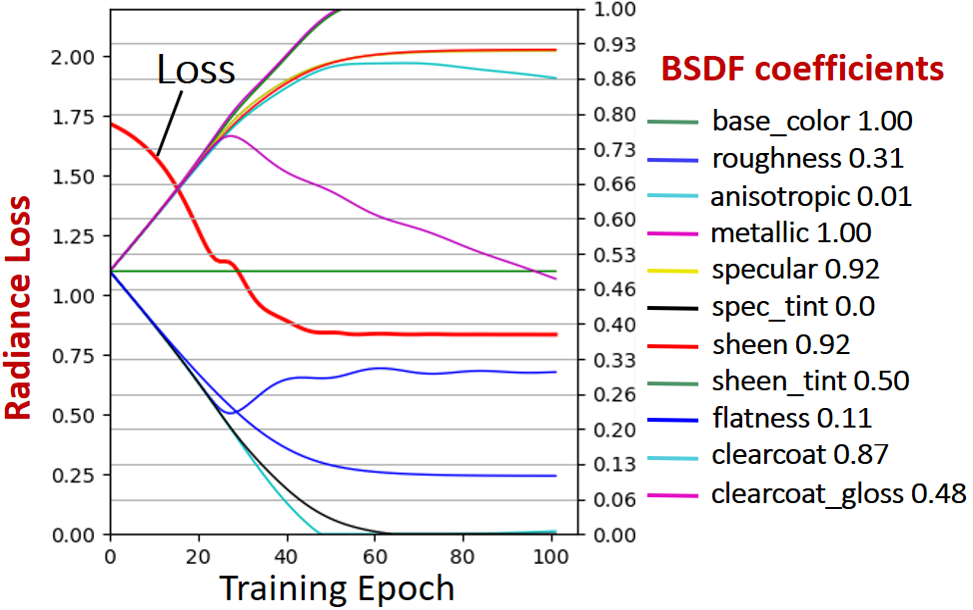}
    \vspace{-1.2\baselineskip}
    \caption{}
    \label{fig_dr2}
\end{subfigure}
\begin{subfigure}{0.20\textwidth}
    \includegraphics[width=\linewidth]{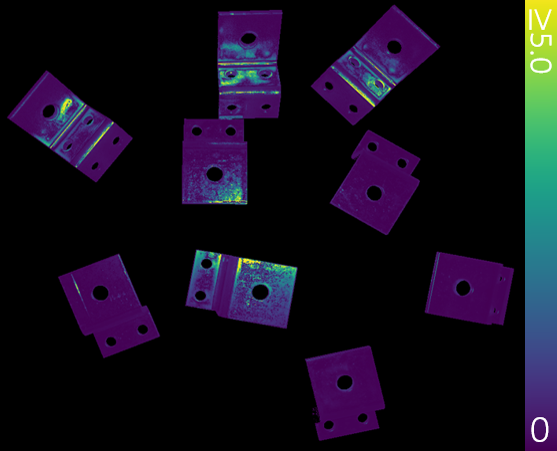}
    \vspace{-1.2\baselineskip}
    \caption{}
    \label{fig_dr3}
\end{subfigure}
\vspace{-0.4\baselineskip}
\caption{(a) Estimating the scene parameters (object BSDF coefficients) with differentiable rendering~\cite{jakob2022mitsuba3}. (b) Loss curve and estimated object BSDF coefficients over the differentiable rendering epochs. (c) Error map between target and estimated radiance when the training converged.}
\label{fig_dr}
\vspace{-0.8\baselineskip}
\end{figure}

\subsection{Recovering Reflection Function}
\label{sec_DR}
The reflection function, $\mathnormal{f}(\boldsymbol{\cdot})$, describes how light interacts with surfaces in the scene. As illustrated in Figure~\ref{fig_imaging}, it takes the physical attributes of a scene (e.g., lighting source, objects' poses, and materials) and outputs the radiance, ${\boldsymbol{E}}$, that reflects into the camera lens. We implement the reflection function, $\mathnormal{f}(\boldsymbol{\cdot})$, as a rendering process that converts the input $\mathbf{x}$ (scene parameters) into the output $\mathbf{y}$ (radiance), and solve this inverse problem using the differentiable rendering technique. The reflection function, $f(\boldsymbol{x})$, is differentiable. Its derivative $\frac{\partial \mathbf{y}}{\partial \mathbf{x}}$ provides a first-order approximation of how a desired output $\mathbf{y}$ (rendered radiance) can be achieved by optimizing the inputs $\mathbf{x}$ (scene parameters). The differentiable loss function, $\mathnormal{l}(\boldsymbol{\mathbf{y}})$, is used to quantify the rendering output $\mathbf{y}$. We demonstrate this process in Figure~\ref{fig_dr1}. The scene parameters (e.g., object materials) can be estimated by minimizing the loss function.

For each object, we estimate its materials with the known projector light, $\boldsymbol{L_p}$, radiance map, $\mathbf{E}_{obj}$ and ground truth 6D object poses, $\mathbf{T}_{c2o}$. The radiance map and object poses are obtained by capturing a static scene of the target objects. We compute the radiance for each pixel of the object surface using the previously recovered photometric response function (Section~\ref{sec_IP}). To acquire object poses, we capture a depth map of the scene and manually label the 6D pose for each object in the camera coordinate. 

We assume the projector light source, $\boldsymbol{L_p}$, is a point light emitter, which radiates the uniform illumination to all directions. For the surface reflection, we use the principled BSDF (bidirectional scattering distribution function)~\cite{burley2015extending} as the surface reflection model. We estimate the BSDF coefficients with the differentiable rendering technique. The optimization problem can be solved with gradient-based methods iteratively. In our approach, we implement the differentiable rendering using the Mitsuba 3 library~\cite{jakob2022mitsuba3}. All parameters are initialized to the medium value and optimized with the L2 loss and the Adam optimizer~\cite{adam}. Figure~\ref{fig_dr2} illustrates the loss curve and estimated BSDF coefficients of a textureless shiny object. The corresponding error map between the target and estimated radiance map is shown in Figure~\ref{fig_dr3}.

\subsection{Predicting Measurement Uncertainty}
\label{sec_pred_unc}
\begin{figure}[t]
\centering
\begin{subfigure}{0.225\textwidth}
    \includegraphics[width=\linewidth]{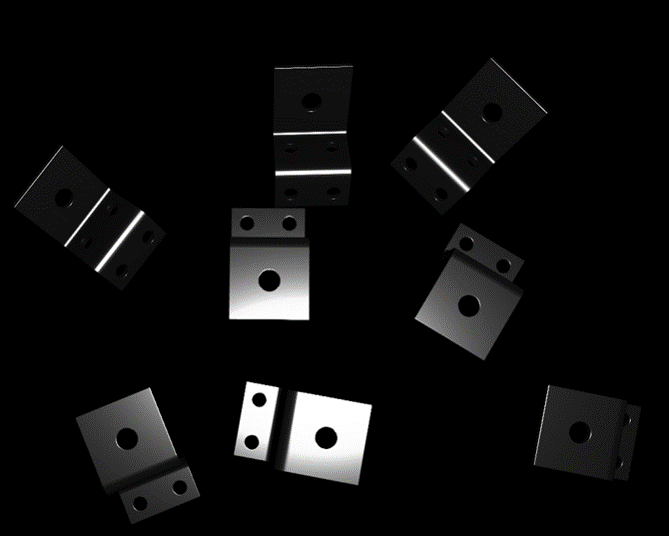}
    \vspace{-1.4\baselineskip}
    \caption{Single-path rendering.}
    \label{fig_single_path}
\end{subfigure}
\begin{subfigure}{0.225\textwidth}
    \includegraphics[width=\linewidth]{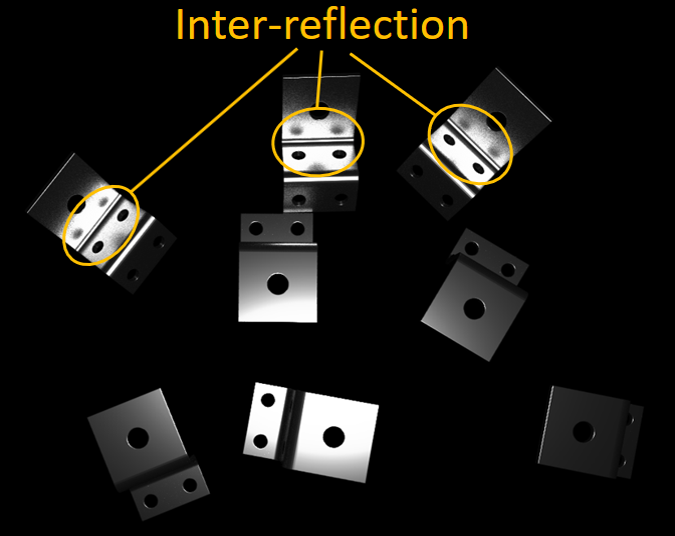}
    \vspace{-1.4\baselineskip}
    \caption{Multi-path rendering.}
    \label{fig_multi_path}
\end{subfigure}
\begin{subfigure}{0.225\textwidth}
    \includegraphics[width=\linewidth]{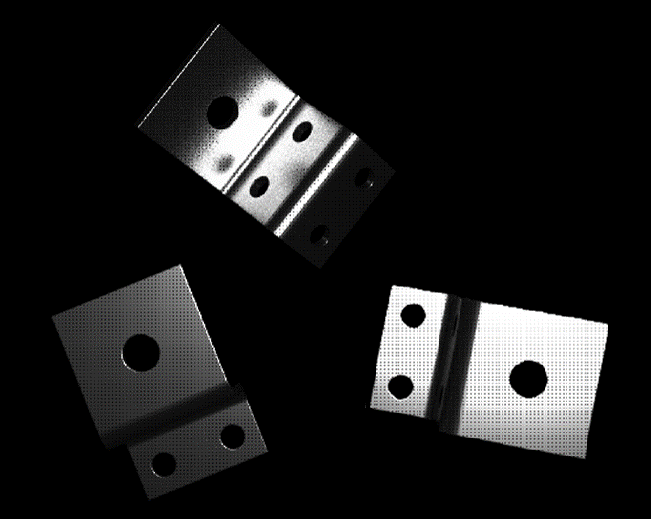}
    \vspace{-1.4\baselineskip}
    \caption{synthesized pattern image.}
    \label{fig_syn_pattern}
\end{subfigure}
\begin{subfigure}{0.225\textwidth}
    \includegraphics[width=\linewidth]{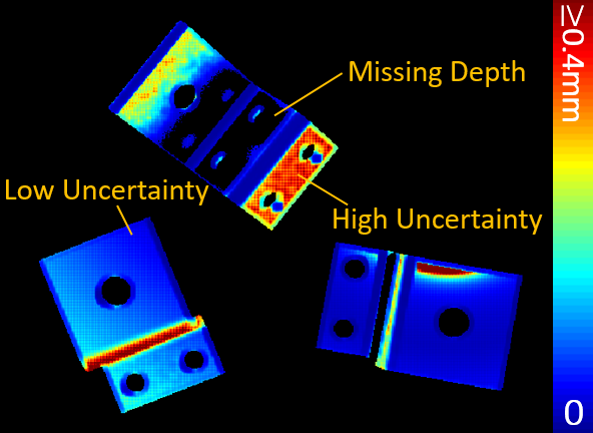}
    \vspace{-1.4\baselineskip}
    \caption{Predicted uncertainty.}
    \label{fig_syn_unc}
\end{subfigure}
\caption{(a)\&(b) Rendering with single- and multi-path ray-tracing, demonstrating the presence of inter-reflection. (c)\&(d) The synthesized pattern projected image of the objects from a future viewpoint and the corresponding predicted depth uncertainty map.}
\label{fig_predict_unc}
\vspace{-0.8\baselineskip}
\end{figure}

For an object, we predict its depth uncertainty, $\check{\boldsymbol{\sigma}}_{z}^2$, from a future camera viewpoint, $\mathbf{T}_{c2w}$, using the forward rendering process. With the recovered reflection function, $\mathnormal{f}\left(\cdot\right)$, and photometric response function, $\mathnormal{g}\left(\cdot\right)$, we can generate a white pattern projected image, $\boldsymbol{I}_w$, of the target object. An object is defined with its CAD model and a 6D pose hypothesis, $\Check{\mathbf{T}}_{w2o}$, defined in the world coordinate:
\begin{equation}
\label{equ_render1}
    \boldsymbol{I}_w = g(\boldsymbol{E} \boldsymbol{\Delta t})
\end{equation}
\begin{equation}
\label{equ_render2}
    \boldsymbol{E} = \mathnormal{f}\left(\boldsymbol{L_p}\:,\:\Check{\mathbf{T}}_{c2o}\right)= \mathnormal{f}\left(\boldsymbol{L_p}\:,\:\mathbf{T}_{c2w}\Check{\mathbf{T}}_{w2o}\right)
\end{equation}
where $\Check{\mathbf{T}}_{c2o}$ is the object pose hypothesis in the camera frame. To predict the missing depth measurement caused by the inter-reflection problem (shown in Figure~\ref{fig_shiny}), we render the object with both multi- and single-path ray tracing. As illustrated in Figure~\ref{fig_single_path}, the rendered image with single-path, $\mathbf{I}_{single}$, only contains direct reflections, which serves as the signal portion to acquire the depth for an SLI camera. The multi-path rendered image (Figure~\ref{fig_multi_path}), $\mathbf{I}_{multi}$, contains both direct and inter-reflections. We treat a pixel depth as missing if the intensity ratio between $\mathbf{I}_{single}$ and $\mathbf{I}_{multi}$ is smaller than a threshold $\tau_I$:
\begin{equation}
\label{equ_set_missing1}
     \Bigl\{\boldsymbol{z}=\emptyset  \:\:\Big|\:\: \forall \:\:  \frac{\mathbf{I}_{single}}{\mathbf{I}_{multi}} < \tau_I\Bigl\}
\end{equation}

To predict the depth uncertainty from the SLI camera, we synthesize a random pattern projected image and compute the uncertainty, $\check{\boldsymbol{\sigma}}_{z}^2$, using Equations~(\ref{equ_ssd})-(\ref{equ_var_depth2}).
We synthesize the random pattern image by combining two multi-path rendered white pattern images with two different lighting intensities (one strong and one weak). Figure~\ref{fig_syn_pattern}-\ref{fig_syn_unc} show an example of our synthesized pattern image and the predicted depth uncertainties. A pixel depth is considered missing when the predicted uncertainty is larger than a pre-defined threshold $\tau_{\sigma}$:
\begin{equation}
\label{equ_set_missing2}
     \Bigl\{\boldsymbol{z}=\emptyset  \:\:\Big|\:\: \forall \: \check{\boldsymbol{\sigma}}_{z} > \tau_{\sigma}\Bigl\}
\end{equation}
Note that, for a candidate viewpoint, a pixel depth measurement is considered to be missing if any condition from Equations~(\ref{equ_set_missing1})-(\ref{equ_set_missing2}) is fulfilled.

\section{Active Pose Refinement with Next-Best-View}
\label{sec_NBV}
In Section~\ref{sec_RF}, we formulate the multi-view object pose refinement problem and solve it using an iterative approach. However, collecting many viewpoints is usually not practical. Hence, in this section, we present an active vision approach for object pose refinement. We developed our NBV policy based on the Fisher information. Compared to most previous NBV approaches~\cite{zhang2018perception,zhang2019beyond}, which neglect the measurement uncertainty, we exploit our predicted depth uncertainties (Section~\ref{sec_pred_unc}) when computing the Fisher information. For each iteration, we estimate the uncertainty of the object pose and find the NBV, which minimizes the predicted uncertainty.

We assume the initial object pose is obtained (e.g., from an external pose estimator) and refine the pose by optimizing the Equation~(\ref{equ_SDF}) with the Jacobian, $\boldsymbol{\mathsf{J}}_{\boldsymbol{\xi}}$, and measurement uncertainty, $\boldsymbol{\Sigma}_{\boldsymbol{\mathsf{sdf}}}$. We compute the covariance of the refined object pose, $\boldsymbol{\Sigma}_{\boldsymbol{\xi}}$, through a first-order approximation of the Fisher information matrix (FIM):
\begin{equation}
\label{equ_cov}
\boldsymbol{\Sigma}_{\boldsymbol{\xi},\mathbf{Z}_{1:K}} = {\left( \boldsymbol{\mathsf{J}}_{\boldsymbol{\xi}, \mathbf{Z}_{1:K}}^T \: \boldsymbol{\Sigma}_{\boldsymbol{\mathsf{sdf}}, \mathbf{Z}_{1:K}}^{-1} \: \boldsymbol{\mathsf{J}}_{\boldsymbol{\xi}, \mathbf{Z}_{1:K}} \right)}^{-1}
\end{equation}
where $\mathbf{Z}_{1:K}$ denote the collected depth measurement sets from $K$ viewpoints. The stacked Jacobian, $\boldsymbol{\mathsf{J}}_{\boldsymbol{\xi}, \mathbf{Z}_{1:K}}$, and measurement uncertainties, $\boldsymbol{\Sigma}_{\boldsymbol{\mathsf{sdf}}, \mathbf{Z}_{1:K}}$, are represented by:
\begin{align}
\label{equ_stack}
    \boldsymbol{\mathsf{J}}_{\boldsymbol{\xi}, \mathbf{Z}_{1:K}} =
    \begin{bmatrix}
    \boldsymbol{\mathsf{J}}_{\boldsymbol{\xi}, \boldsymbol{Z}_{1}} \\
    \vdots \\
    \boldsymbol{\mathsf{J}}_{\boldsymbol{\xi}, \boldsymbol{Z}_{K}}
    \end{bmatrix}
    \:,\:
    \boldsymbol{\Sigma}_{\boldsymbol{\mathsf{sdf}}, \mathbf{Z}_{1:K}} =
    \begin{bmatrix}
    \boldsymbol{\Sigma}_{\boldsymbol{\mathsf{sdf}}, \mathbf{Z}_1} & {} & {}\\
    {} & \ddots & {}\\
    {} & {} & \boldsymbol{\Sigma}_{\boldsymbol{\mathsf{sdf}}, \mathbf{Z}_K}\\
    \end{bmatrix}  
\end{align}
The row-blocks, $\boldsymbol{\mathsf{J}}_{\boldsymbol{\xi}, \boldsymbol{Z}_{k}}$ and $\boldsymbol{\Sigma}_{\boldsymbol{\mathsf{sdf}}, \mathbf{Z}_k}$, correspond to the Jacobian matrix and $\boldsymbol{\mathsf{SDF}}$ uncertainty with the $k^{th}$ viewpoint, and can be calculated using Equation~(\ref{equ_jacobian_Tow}) and (\ref{equ_propagate}), respectively. To compute the uncertainty for the object pose covariance, we use the differential entropy, $h_e\left( \boldsymbol{\Sigma}_{\boldsymbol{\xi},\mathbf{Z}_{1:K}} \right)$:
\begin{align}
\label{equ_differential_entropy}
h_e\left( \boldsymbol{\Sigma}_{\boldsymbol{\xi},\mathbf{Z}_{1:K}} \right) = \frac{1}{2} \ln{\left(\left(2\pi e\right)^n \left| \boldsymbol{\Sigma}_{\boldsymbol{\xi}, \boldsymbol{Z}_{1:K}} \right| \right)}
\end{align}
where $h_e\left( \boldsymbol{\Sigma}_{\boldsymbol{\xi},\mathbf{Z}_{1:K}} \right)$ is expressed in nats.

To increase the object pose accuracy, we aim to find the next best camera viewpoint $\boldsymbol{v}^*$ from a set of candidate viewpoints $\boldsymbol{\{V\}}$ which will minimize the entropy of
the object pose, $h_e\left( \boldsymbol{\Sigma}_{\boldsymbol{\xi}} \right)$. Suppose we have collected the depth measurement sets, $\mathbf{Z}_{1:K}$, from $K$ viewpoints. For a future camera viewpoint, $\widehat{\boldsymbol{v}}$, the stacked Jacobian and measurement uncertainties have the following form:
\begin{align}
\label{equ_jacobian_NBV}
    \boldsymbol{\mathsf{J}}_{\boldsymbol{\xi}, \overline{\boldsymbol{Z}}} =
    \begin{bmatrix}
    \boldsymbol{\mathsf{J}}_{\boldsymbol{\xi}, \boldsymbol{Z}_{1:K}} \\
    \boldsymbol{\mathsf{J}}_{\boldsymbol{\xi}, \widehat{\boldsymbol{Z}}}
    \end{bmatrix}    
    \:\:,\:\:
    \boldsymbol{\Sigma}_{\boldsymbol{\mathsf{sdf}}, \overline{\boldsymbol{Z}}} =
    \begin{bmatrix}
    \boldsymbol{\Sigma}_{\boldsymbol{\mathsf{sdf}}, \mathbf{Z}_{1:K}} & {\mathbf{0}}\\
    {\mathbf{0}} & \boldsymbol{\Sigma}_{\boldsymbol{\mathsf{sdf}}, \widehat{\boldsymbol{Z}}}\\
    \end{bmatrix}      
\end{align}
where $\overline{\boldsymbol{Z}} = \bigl\{ \boldsymbol{Z}_{1:K}, \widehat{\boldsymbol{Z}} \bigl\}$ includes acquired measurement sets $\boldsymbol{Z}_{1:K}$ from viewpoints $\boldsymbol{v}_{1:K}$ and predicted measurement set $\widehat{\boldsymbol{Z}}$ for the future viewpoint, $\widehat{\boldsymbol{v}}$. With the FIM evaluation, we can predict the object pose covariance by:
\begin{equation}
\label{equ_cov_nbv}    \boldsymbol{\Sigma}_{\boldsymbol{\xi},\overline{\boldsymbol{Z}}} = {\left( \boldsymbol{\mathsf{J}}_{\boldsymbol{\xi}, \overline{\boldsymbol{Z}}}^T \:\: \boldsymbol{\Sigma}_{\boldsymbol{\mathsf{sdf}}, \overline{\boldsymbol{Z}}}^{-1} \:\: \boldsymbol{\mathsf{J}}_{\boldsymbol{\xi}, \overline{\boldsymbol{Z}}} \right)}^{-1}
\end{equation}
Note that, in Equation~(\ref{equ_jacobian_NBV}), we compute the Jacobian $\boldsymbol{\mathsf{J}}_{\boldsymbol{\xi}, \widehat{\boldsymbol{Z}}}$ and uncertainty $\boldsymbol{\Sigma}_{\boldsymbol{\mathsf{sdf}}, \widehat{\boldsymbol{Z}}}$ before actually moving to the camera viewpoint $\widehat{\boldsymbol{v}}$. The computation of Jacobian $\boldsymbol{\mathsf{J}}_{\boldsymbol{\xi}, \widehat{\boldsymbol{Z}}}$ is based on the initial object pose guess. We compute the $\boldsymbol{\mathsf{SDF}}$ uncertainty, $\boldsymbol{\Sigma}_{\boldsymbol{\mathsf{sdf}}, \widehat{\boldsymbol{Z}}}$, using the online rendering process (described in Section~\ref{sec_pred_unc}).

Our NBV is determined over candidate viewpoints $\boldsymbol{\{V\}}$ by minimizing the predicted entropy of the object pose:
\begin{align}
\label{equ_IG_max}
\boldsymbol{v}^{*} = \argmin_{\widehat{\boldsymbol{v}}} \:\: {h_e}\left(\boldsymbol{\Sigma}_{\boldsymbol{\xi},\overline{\boldsymbol{Z}}}\right)
\end{align}
Once the next-best-view $\boldsymbol{v}^{*}$ is determined, the camera is moved, and a measurement set $\boldsymbol{Z}^*$ is collected from the corresponding viewpoint. We append the measurement set by $\mathbf{Z}_{1:K} \cup \mathbf{Z}^{*} \rightarrow \mathbf{Z}_{1:K+1}$ to recompute the object pose and perform the NBV selection again using Equations~(\ref{equ_jacobian_NBV})-(\ref{equ_IG_max}). This process is repeated until the predicted entropy falls below a user-defined threshold or a maximum number of views is selected.

\begin{table*}[t]
\resizebox{\textwidth}{!}{
\begin{tabular}{|cc||cc||cc||cc||cc||cc||cc|}
\hline
\multicolumn{2}{|c||}{\multirow{2}{*}{\backslashbox{Method}{Objects}}}         & \multicolumn{2}{c||}{Eye Bolt}                      & \multicolumn{2}{c||}{Tube Fitting}                  & \multicolumn{2}{c||}{Chrome Screw}                  & \multicolumn{2}{c||}{Gear}                          & \multicolumn{2}{c||}{Zigzag}                        & \multicolumn{2}{c|}{ALL}                           \\ \cline{3-14} 
\multicolumn{2}{|c||}{}                                & \multicolumn{1}{c|}{(5, 5)}        & (2, 2)        & \multicolumn{1}{c|}{(5, 5)}        & (2, 2)        & \multicolumn{1}{c|}{(5, 5)}        & (2, 2)        & \multicolumn{1}{c|}{(5, 5)}        & (2, 2)        & \multicolumn{1}{c|}{(5, 5)}        & (2, 2)        & \multicolumn{1}{c|}{(5, 5)}        & (2, 2)        \\ \hline
\multicolumn{2}{|c||}{Initial Pose}                    & \multicolumn{1}{c|}{9.3}           & 0.19          & \multicolumn{1}{c|}{18.4}          & 0.87          & \multicolumn{1}{c|}{34.5}          & 5.3           & \multicolumn{1}{c|}{24.6}          & 1.3           & \multicolumn{1}{c|}{19.9}          & 1.14          & \multicolumn{1}{c|}{21.3}          & 1.77          \\ \hline
\multicolumn{1}{|c|}{\multirow{2}{*}{1 View}}  & Ours & \multicolumn{1}{c|}{\textbf{87.2}} & 55.1          & \multicolumn{1}{c|}{\textbf{69.4}} & \textbf{42.3} & \multicolumn{1}{c|}{54.2}          & 13.5          & \multicolumn{1}{c|}{\textbf{80.4}} & \textbf{70.7} & \multicolumn{1}{c|}{\textbf{96.9}} & \textbf{87.3} & \multicolumn{1}{c|}{\textbf{77.6}} & \textbf{53.8} \\ \cline{2-14} 
\multicolumn{1}{|c|}{}                         & ICP  & \multicolumn{1}{c|}{81.3}          & \textbf{56.2} & \multicolumn{1}{c|}{68.3}          & 35.5          & \multicolumn{1}{c|}{\textbf{60.1}} & \textbf{13.7} & \multicolumn{1}{c|}{79.8}          & 66.2          & \multicolumn{1}{c|}{96.0}          & 74.7          & \multicolumn{1}{c|}{77.1}          & 49.3          \\ \hline
\multicolumn{1}{|c|}{\multirow{2}{*}{2 Views}} & Ours & \multicolumn{1}{c|}{\textbf{91.8}} & \textbf{71.0} & \multicolumn{1}{c|}{\textbf{83.2}} & \textbf{61.0} & \multicolumn{1}{c|}{69.6}          & 17.5          & \multicolumn{1}{c|}{\textbf{92.5}} & \textbf{88.8} & \multicolumn{1}{c|}{\textbf{96.8}} & \textbf{87.8} & \multicolumn{1}{c|}{\textbf{86.8}} & \textbf{65.2} \\ \cline{2-14} 
\multicolumn{1}{|c|}{}                         & ICP  & \multicolumn{1}{c|}{87.4}          & 67.8          & \multicolumn{1}{c|}{76.9}          & 44.5          & \multicolumn{1}{c|}{\textbf{73.8}} & \textbf{18.1} & \multicolumn{1}{c|}{91.4}          & 85.2          & \multicolumn{1}{c|}{96.2}          & 82.9          & \multicolumn{1}{c|}{85.1}          & 59.7          \\ \hline
\multicolumn{1}{|c|}{\multirow{2}{*}{4 Views}} & Ours & \multicolumn{1}{c|}{\textbf{94.1}} & \textbf{77.7} & \multicolumn{1}{c|}{\textbf{89.8}} & \textbf{75.0} & \multicolumn{1}{c|}{76.2}          & 19.6          & \multicolumn{1}{c|}{\textbf{97.9}} & \textbf{96.4} & \multicolumn{1}{c|}{\textbf{96.8}} & \textbf{88.4} & \multicolumn{1}{c|}{\textbf{91.0}} & \textbf{71.4} \\ \cline{2-14} 
\multicolumn{1}{|c|}{}                         & ICP  & \multicolumn{1}{c|}{88.6}          & 74.2          & \multicolumn{1}{c|}{82.8}          & 53.6          & \multicolumn{1}{c|}{\textbf{77.7}} & \textbf{21.2} & \multicolumn{1}{c|}{96.4}          & 93.1          & \multicolumn{1}{c|}{96.0}          & 83.7          & \multicolumn{1}{c|}{88.3}          & 65.2          \\ \hline
\end{tabular}}
\vspace{-0.5\baselineskip}
\caption{Object pose refinement results on the ENSENSO test set from the ROBI dataset~\cite{yang2021robi}, expressed as the correct detection rate. Both ICP and our pose refinement module are provided the same depth data from the same viewpoint(s). An object pose is considered correct if it lies within 5-mm/5-degree (5, 5), or 2-mm/2-degree (2, 2), of ground truth.}
\vspace{-1.0\baselineskip}
\label{tab_refinement}
\end{table*}

\begin{figure*}[t]
\centering
\begin{subfigure}{0.19\textwidth}
  \includegraphics[width=\linewidth]{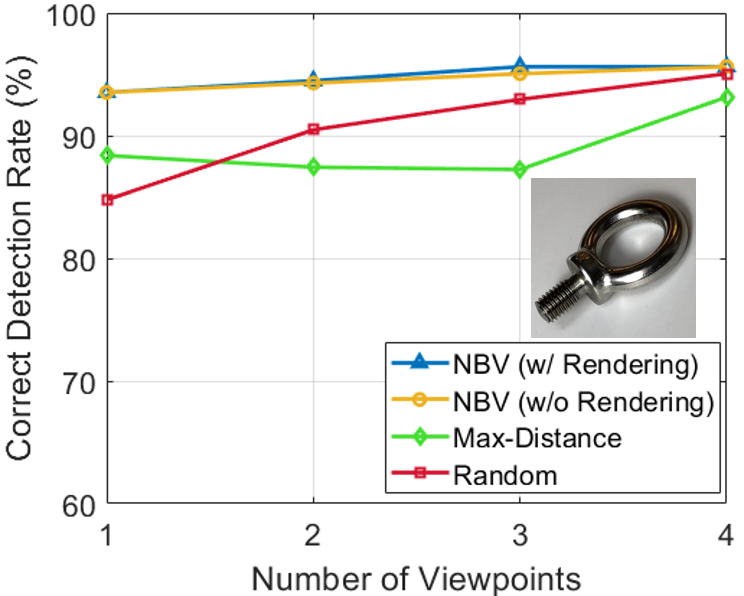}
  \caption{Eye Bolt}  
  \vspace{-0.4\baselineskip}
\end{subfigure}
\begin{subfigure}{0.19\textwidth}
  \includegraphics[width=\linewidth]{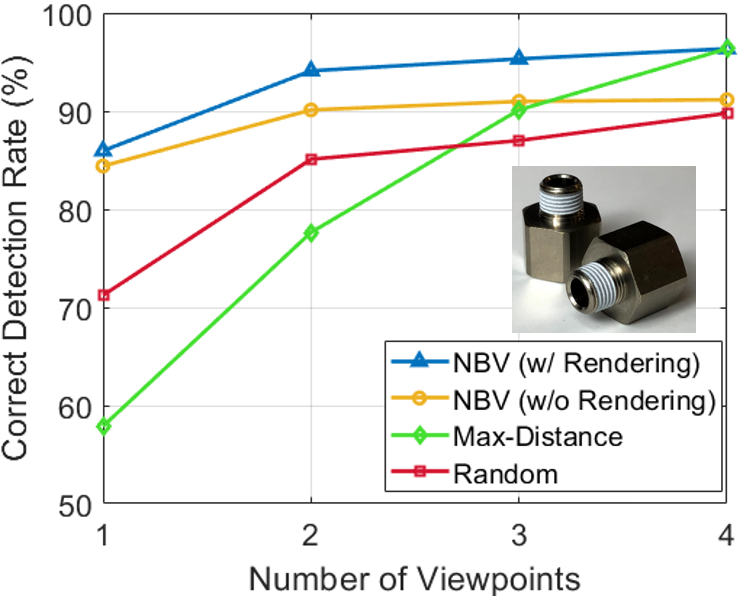}
  \caption{Tube Fitting}  
  \vspace{-0.4\baselineskip}
\end{subfigure}
\begin{subfigure}{0.19\textwidth}
  \includegraphics[width=\linewidth]{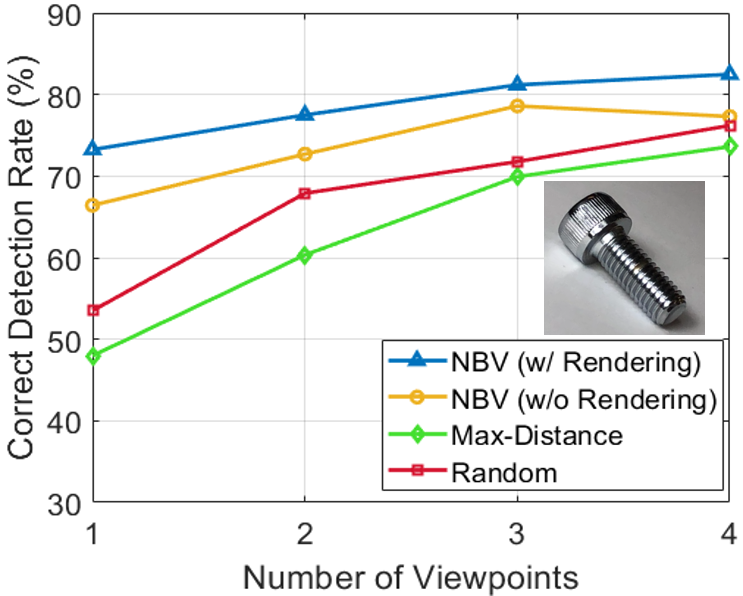}
  \caption{Chrome Screw}  
  \vspace{-0.4\baselineskip}
\end{subfigure}
\begin{subfigure}{0.19\textwidth}
  \includegraphics[width=\linewidth]{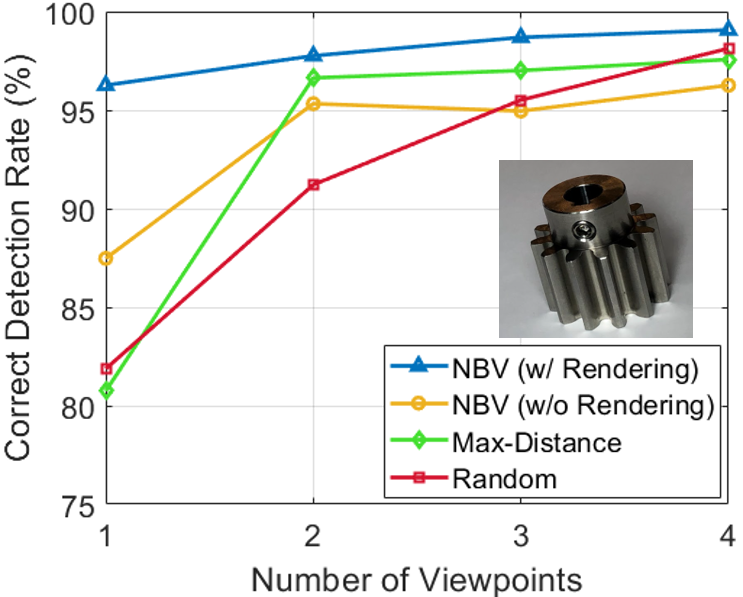}
  \caption{Gear}
  \vspace{-0.4\baselineskip}
  \label{fig_cov}
\end{subfigure}
\begin{subfigure}{0.19\textwidth}
  \includegraphics[width=\linewidth]{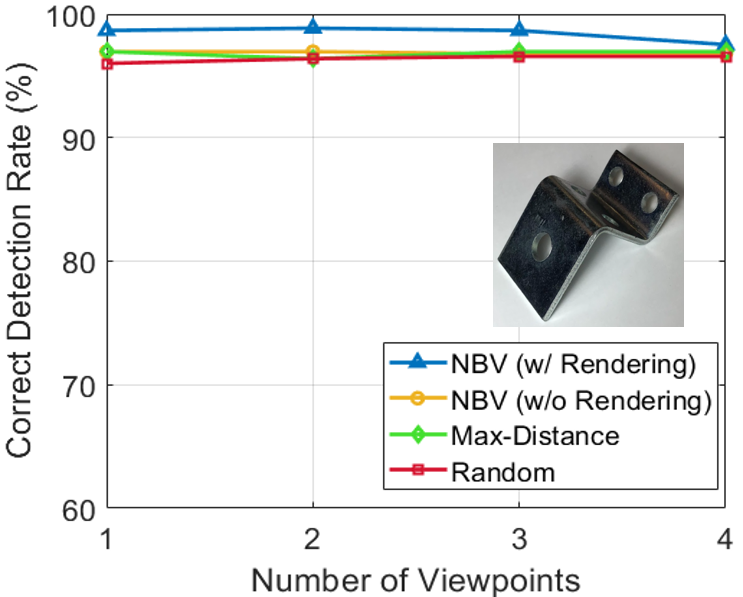}
  \caption{Zigzag}
  \vspace{-0.4\baselineskip}
  \label{fig_cov}
\end{subfigure}
\caption{Evaluation of our next-best-view policy when comparing against two heuristic-based baselines. We use our pose refinement module for all the viewpoint selection strategies. The results are evaluated using the correct detection rate with the 5-mm/5-degree metric. Our approach can achieve a high correct detection rate with much fewer viewpoints.}
\vspace{-1.2\baselineskip}
\label{fig_nbv_result}
\end{figure*}

\section{EXPERIMENTS}
\label{sec_EXP}
To show the advantage of our active pose refinement system, we want to answer two questions: (1) Can our pose refinement module recover accurate object poses given depth measurements? (2) Can our active vision policy achieve optimal performance with a minimal number of viewpoints? To answer the question (1), we compare our pose refinement module with the classical ICP algorithm, given the same input depth data. For question (2), we use our pose refinement module and test our active vision approach against two heuristic-based policies.

\subsection{Datasets and Evaluation Metrics}
In our experiments, we use an industrial-grade SLI camera (IDS ENSNESO N35), which equips with two cameras and a visible-light projector. We evaluate our method on the ROBI dataset~\cite{yang2021robi}, which was captured using this camera. The ROBI dataset provides multi-view depth maps and pattern-projected images for shiny objects. The precisely labelled ground truth 6D object poses are also provided. In our experiments, we pick five objects that are the most shiny and evaluate each object individually.

We evaluate the 6D object pose accuracy using the correct detection rate with the 5-mm/5-degree (5,5) and 2-mm/2-degree (2,2) metrics. The 5-mm/5-degree metric considers a refined pose correct if the translation error is smaller than 5 mm and the rotation error is smaller than 5 degrees. In our evaluation, a ground truth pose will be considered if its visibility is larger than 80\%.

\subsection{Object Pose Refinement Evaluation}
\begin{figure}[t]
\centering
\begin{subfigure}{0.157\textwidth}
  \includegraphics[width=\linewidth]{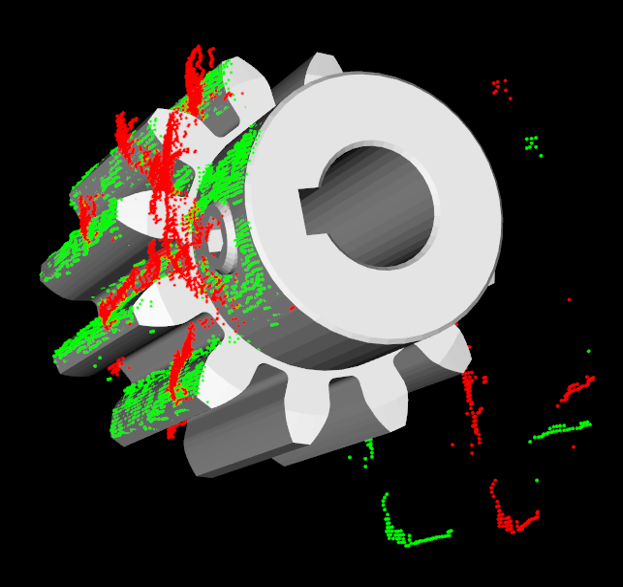}
  \vspace{-1.4\baselineskip}
  \caption{Gear}
  \vspace{-0.4\baselineskip}
  \label{fig_refine_GEAKS}
\end{subfigure}
\begin{subfigure}{0.157\textwidth}
    \includegraphics[width=\linewidth]{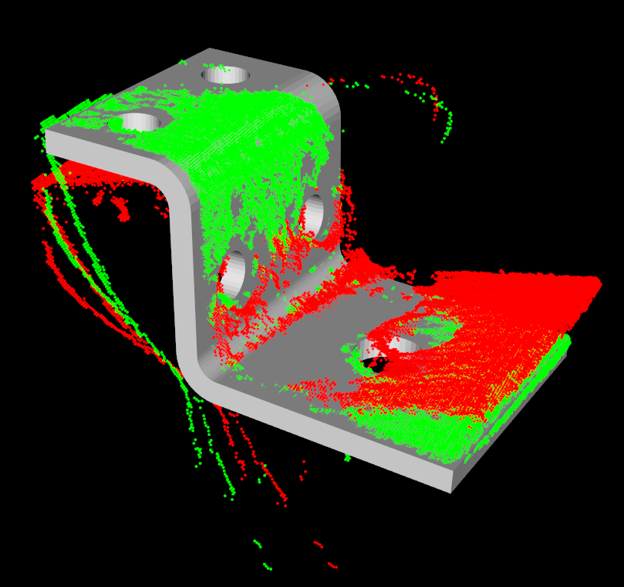}
    \vspace{-1.4\baselineskip}
    \caption{Zigzag}
    \vspace{-0.4\baselineskip}
    \label{fig_refine_ZIGZAG}
\end{subfigure}
\begin{subfigure}{0.157\textwidth}
    \includegraphics[width=\linewidth]{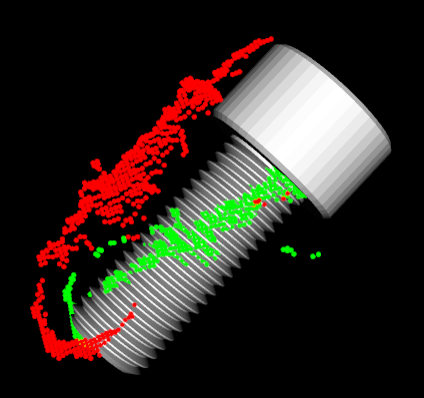}
    \vspace{-1.4\baselineskip}
    \caption{Chrome Screw}
    \vspace{-0.4\baselineskip}
    \label{fig_refine_SCREW}
\end{subfigure}
\caption{Example refinement results on the ROBI dataset. The red and green point clouds are transformed by initial and refined object pose, respectively.}
\vspace{-1.4\baselineskip}
\label{fig_refine}
\end{figure}

We first visualize how the poses of objects are refined with our refinement approach in Figure~\ref{fig_refine}. It can be seen that, our pose refinement module can achieve a reliable and accurate refinement result even in the presence of many outliers and the initial pose has a large error.

For quantitative evaluation, we compare our proposed refinement approach against the widely-used ICP algorithm with the same input depth data. We perform the pose evaluation on each individual object. Specifically, we first utilize the instance segmentation network from~\cite{yang20236d} to segment the objects in the scene. The segmented object is then fed into a template matching-based pose estimation approach, Line-2D~\cite{hinterstoisser2011gradient}, to acquire the initial object pose guesses. An initial object pose will be used for the refinement evaluation if its pose error satisfies the 30-mm/30-degree metric. For each object, we evaluate the refinement accuracy with 1, 2, and 4 camera viewpoints, which are selected randomly. We apply the object segmentation mask on the depth map from each viewpoint to acquire the object's depth measurements. The same initial object poses and depth measurements are fed into these two pose refinement methods for evaluation.

The results of the object refinement are summarized in Table~\ref{tab_refinement}. We can see that, the poses acquired with only Line-2D pose estimator have low detection rates for both 5-mm/5-degree and 2-mm/2-degree metrics. The results can be significantly improved when performing the pose refinement with the depth data. Compared to the ICP algorithm, our refinement approach outperforms it in almost all tests. With the 5-mm/5-degree metric, our pose refinement module outperforms ICP by 0.5\% for 1-view, 1.7\% for 2-view, and 2.7\% for 4-view test set. When using a more strict 2-mm/2-degree metric, our refinement method outperforms ICP by a more significant margin of 4.5\%, 5.5\%, and 6.2\%  on the 1-view, 2-view, and 4-view test sets, respectively. However, it is noteworthy that our refinement approach performs worse than ICP on the object "Chrome Screw", especially on the 1-view test set. This is because the object is extremely shiny and has many missing depth measurements on the surface. Moreover, as illustrated in Figure~\ref{fig_refine_SCREW}, this object lacks geometric constraints due to its cylindrical shape, making it difficult for the optimization to find the global minima. Hence, selecting informative viewpoints to acquire sufficient depth measurements is crucial.

\subsection{Next-Best-View Evaluation}
We compare our next-best-view approach against two heuristic-based strategies as the baselines. The first baseline, "Random", selects viewpoints randomly from the candidate viewpoints. The second baseline, "Max-Distance" moves the camera to the furthest distance location from previous viewpoints. For all view selection strategies, we use the same pose refinement module (our SDF-based approach) for a fair comparison.

Figure~\ref{fig_nbv_result} presents the results when using the 5-mm/5-degree metric. We can see that, our NBV policy (blue curve) outperforms the two baselines (red and green curves) by a large margin. To achieve the same level of correct detection rate, our proposed NBV approach requires much fewer viewpoints than the baselines. This phenomenon is more obvious when using fewer and fewer viewpoints. When compared to the "Random" strategy, our NBV approach outperforms it by 12.1\% for 1-view and 6.3\% for 2-view test set. Compared to the "Max-Distance", the NBV policy exceeds it by 15.1\% and 8.8\% for 1-view and 2-view, respectively. For the shiniest object "Chrome Screw", our NBV policy can achieve a high detection rate, 73.2\%, with only one viewpoint. This result is comparable to the baseline policies when using four views (76.2\% for the "Random", 73.6\% for the "Max-Distance"). This is particularly valuable for applications that have strict cycle time requirement, such as robotic bin-picking.

As presented in Section~\ref{sec_Predict_depth_uc} and \ref{sec_NBV}, a key component of our NBV policy is the depth uncertainty prediction of future viewpoints by online rendering. To demonstrate its effectiveness, we implement an alternative version of our NBV approach, one which assumes
the depth uncertainty is constant for different future camera viewpoints. This version does not require online rendering and predicts the object pose covariance (Equation~\ref{equ_cov_nbv}) with the Jacobian approximation only. As shown in Figure~\ref{fig_nbv_result}, the NBV can achieve high performance without predicting the depth uncertainty (yellow curve) for the object "Eye Bolt" and "Zigzag". This is because these two objects have low specular reflection, and the depth uncertainty is consistent for a wide range of different viewpoints. However, when objects have strong specular reflection (e.g., "Chrome Screw", "Tube Fitting", "Gear"), the NBV performance can be significantly improved by including the depth uncertainty prediction module.

\section{CONCLUSIONS AND FUTURE WORK}
In this paper, we present a complete active vision framework of 6D pose refinement and next-best-view prediction for shiny objects. Based on the SLI camera, we first estimate the uncertainty of depth measurements and integrate them into our object pose refinement module. Our framework refines the object pose and selects the next-best-view by minimizing the predicted uncertainty. We evaluate our approach on a challenging real-world dataset that includes the shiny objects captured from multiple viewpoints. The results demonstrate that our pose refinement module outperforms the classical ICP algorithm when using the same input depth data, and our NBV policy can achieve high pose refinement accuracy with significantly fewer viewpoints when compared to heuristic baselines. In future work, we will investigate how to include the initial object pose estimation into our active vision framework, and explore how RGB images can be leveraged in a similar way, eliminating the specialization of our approach to the SLI camera setting.






\bibliographystyle{ieeetr}
\bibliography{bibliography.bib}

\end{document}